%% file: acl_latex.tex
\pdfoutput=1
\documentclass[11pt]{article}

\usepackage[final]{acl}

\usepackage{times}
\usepackage{latexsym}

\usepackage[T1]{fontenc}
\usepackage{array}
\usepackage[utf8]{inputenc}

\usepackage{microtype}

\usepackage{inconsolata}
\usepackage{CJKutf8}
\usepackage{graphicx}

\usepackage{color, xcolor}

\usepackage[utf8]{inputenc}
\usepackage{booktabs}
\usepackage{algorithm}
\usepackage{algorithmic}
\usepackage{multicol}
\usepackage{multirow, makecell, caption}
\usepackage{colortbl}
\usepackage{tikz}
\usepackage{arydshln}
\usepackage{pgfplots}
\usepackage{amsmath}
\usepackage{amssymb}
\usepackage{tcolorbox}
\usepackage{tabularx}
\usepackage{enumerate}
\usepackage{pifont}
\makeatletter
\def\adl@drawiv#1#2#3{%
        \hskip.5\tabcolsep
        \xleaders#3{#2.5\@tempdimb #1{1}#2.5\@tempdimb}%
                #2\z@ plus1fil minus1fil\relax
        \hskip.5\tabcolsep}
\newcommand{\cdashlinelr}[1]{%
  \noalign{\vskip\aboverulesep
           \global\let\@dashdrawstore\adl@draw
           \global\let\adl@draw\adl@drawiv}
  \cdashline{#1}
  \noalign{\global\let\adl@draw\@dashdrawstore
           \vskip\belowrulesep}}
\makeatother

\title{
ChemAmp: Amplified Chemistry Tools via Composable Agents}

\author{
\textbf{Zhucong Li\textsuperscript{1,3}\thanks{\enspace Equal contribution.}},
 \textbf{Powei Chang\textsuperscript{2}\footnotemark[1]},
 \textbf{Jin Xiao\textsuperscript{2}},
  \textbf{Zhijian Zhou\textsuperscript{1}},
 \\
 \textbf{Qianyu He\textsuperscript{3}},
  \textbf{Jiaqing Liang\textsuperscript{2}},
\textbf{Fenglei Cao\textsuperscript{4}},
\textbf{Xu Yinghui\textsuperscript{1}},
  \textbf{Yuan Qi\textsuperscript{1,4,5}\thanks{\enspace Corresponding author.}}
  \\
  \textsuperscript{1}Artificial Intelligence Innovation and Incubation Institute, Fudan University, 
 \\
  \textsuperscript{2}School of Data Science, Fudan University, 
  \\
 \textsuperscript{3}College of Computer Science and Artificial Intelligence, Fudan University, 
 \\
 \textsuperscript{4}Shanghai Academy of Artificial Intelligence for Science, 
 \\
 \textsuperscript{5}Department of Information and Intelligence Development, Zhongshan Hospital, Fudan University
 \\
    \{zcli22, bwzhang24, jinxiao23\}@m.fudan.edu.cn,
    \{liangjiaqing, xuyinghui, qiyuan\}@fudan.edu.cn
}

\begin{document}
\thispagestyle{empty}
\maketitle

\begin{abstract}
\input{00abstract}

\end{abstract}

\section{Introduction}
\input{01intro}

\section{Related Work}
\input{02related}

\section{ChemAmp Framework}
\input{04method}

\section{Experiments}
\input{05experiment}

\section{Analysis}
\input{ana}

\section{Conclusion}
\input{06conclusion}

\bibliography{custom}
\clearpage

\appendix
\renewcommand{\thesection}{\Alph{section}} 
\section*{Appendix}
\input{Appendix}

\end{document}

%% file: 00abstract.tex
Although LLM-based agents are proven to master tool orchestration in scientific fields, particularly chemistry, their single-task performance remains limited by underlying tool constraints.
To this end, we propose tool amplification, a novel paradigm that enhances the collective capabilities of specialized tools through optimized, dynamic coordination within individual tasks.
Instantiating this paradigm, we introduce ChemAmp, a computationally lightweight framework that dynamically treats chemistry tools (e.g., UniMol2, Chemformer) as composable building-block agents. It constructs task-specialized super-agents that transcend atomic tool constraints with limited data ($\leq$10 samples). Our evaluations across four core chemistry tasks molecular design, molecule captioning, reaction prediction, and property prediction demonstrate that ChemAmp outperforms chemistry-specialized models, generalist LLMs, and agent systems with tool orchestration. Critically, this bottom-up construction strategy enables 94\% inference token cost reductions versus vanilla multi-agent systems.
Our code and dataset are publicly available at 
\url{https://github.com/Chang-pw/ChemAmp}.

%% file: 01intro.tex
\input{figures/intro_fig}

Large Language Model (LLM)-based agents ~\citep{touvron2023llama,achiam2023gpt,deepseekai2025deepseekr1incentivizingreasoningcapability} have emerged as powerful tools for automating complex scientific workflows, particularly in chemistry, where they orchestrate multi-step processes involving specialized computational tools ~\citep{bran2023chemcrowaugmentinglargelanguagemodels,ouyang2024structuredchemistryreasoninglarge,boiko2023autonomous, han2024generalistspecialistsurveylarge, shi2023relmleveraginglanguagemodels, yu2025chemtoolagentimpacttoolslanguage, tang2025chemagentselfupdatinglibrarylarge}.

As shown in Fig \ref{fig:intro} (Left), while effective for task orchestration, current works primarily sequence pre-defined tool usage. 
For example, when an LLM-based agent handles three potential pipeline tasks such as molecular design~\citep{noutahi2023gottasafenewframework}, reaction prediction~\citep{shi2023relmleveraginglanguagemodels}, and property prediction~\citep{srinivas2024crossmodallearningchemistryproperty}, it sequentially selects the predefined chemistry tools for each task to obtain computational results. 
However, a critical limitation remains unaddressed: the performance of these agents on individual, complex tasks is fundamentally constrained by the inherent capabilities and scope of the underlying tools they invoke. As a result, errors can propagate through the reasoning chain. This bottleneck restricts performance gains and often leads to inefficient, redundant tool calls that inflate computational costs.

In this work, we confront a more fundamental limitation: even state-of-the-art chemistry tools (e.g., UniMol2, Chemformer, ChemDFM)~\citep{ji2024uni, irwin_chemformer_2022,zhao2024chemdfmlargelanguagefoundation} remain constrained by their atomic capabilities when operating in isolation. As illustrated in Fig. \ref{fig:intro} (Right), we posit a fundamental shift from tool orchestration to tool amplification where tools transcend native capabilities through hierarchical coordination within atomic tasks. Unlike traditional tool orchestration (which schedules tools across different tasks), amplification focuses on making tools work better together within a single task. By dynamically combining tools into collaborative teams, they can achieve capabilities beyond what any tool can do alone.

This amplification paradigm introduces two core challenges:
(1) Structural Search Complexity: Optimal agent-tool compositions vary significantly across chemistry tasks due to domain-specific constraints.
(2) Efficiency-Capability Tradeoff: Manual composition is infeasible, while exhaustive search incurs prohibitive computational costs.

To address these challenges, we introduce \textbf{ChemAmp} (\textbf{Chem}istry Tool \textbf{Amp}lification), a framework realizing tool amplification through composable agents. Departing from naive stacking, 
ChemAmp employs a \textbf{bi-phase encapsulation engine} constructing agent hierarchies from atomic tools via bottom-up iterative composition.
Specifically, ChemAmp operates through two synergistic stages:
(1) \textbf{Atomic-to-Composite Amplification}: 
    Atomic tools undergo iterative encapsulation into sub-agents, 
    evolving optimal combinations through adaptive scoring and automated feedback.
(2) \textbf{Cross-Composite Synergy}: 
    Sub-agents are further encapsulated into composite networks through the same mechanism, 
    amplifying capabilities via emergent hierarchical coordination.

We deploy ChemAmp across four foundational chemistry domains: molecular design, molecule captioning, property prediction, and reaction prediction. For each task, ChemAmp discovers optimal agent compositions, dynamically constructing task-specialized super-agents that transcend atomic tool constraints with limited data($\leq10$ samples).
Experimental results demonstrate the consistent superiority of ChemAmp over three critical baselines: chemistry-specialized ~\citep{ji2024unimol2exploringmolecularpretraining,zhao2024chemdfmlargelanguagefoundation}, generalist LLMs~\citep{openai2024gpt4ocard,deepseekai2025deepseekr1incentivizingreasoningcapability}, and agent systems with tool orchestration\citep{bran2023chemcrowaugmentinglargelanguagemodels, yu2025chemtoolagentimpacttoolslanguage}. Notably, ChemAmp delivers significant gains while requiring only 6\% inference token costs versus basic multi-agent systems, validating both the efficacy and efficiency of tool amplification.
This paradigm shift enables ChemAmp to automatically construct task-specialized super-agents that amplify tool capabilities.

To summarize, our contributions are three-fold:  
\begin{itemize}  
    \item We propose tool amplification, a novel paradigm that enhances the collective capabilities of specialized tools through optimized, dynamic coordination within individual tasks.
    \item We develop ChemAmp, a lightweight framework realizing amplification via bi-phase iterative encapsulation of atomic tools into agent composite tools.  
    \item We demonstrate state-of-the-art results across molecular design, captioning, reaction prediction, property prediction and 94\% inference token cost reductions versus vanilla multi-agent systems.  
\end{itemize}
\input{figures/method_fig}

%% file: figures/intro_fig.tex
\begin{figure*}[!ht] 
    \centering
        \includegraphics[width=1\textwidth]{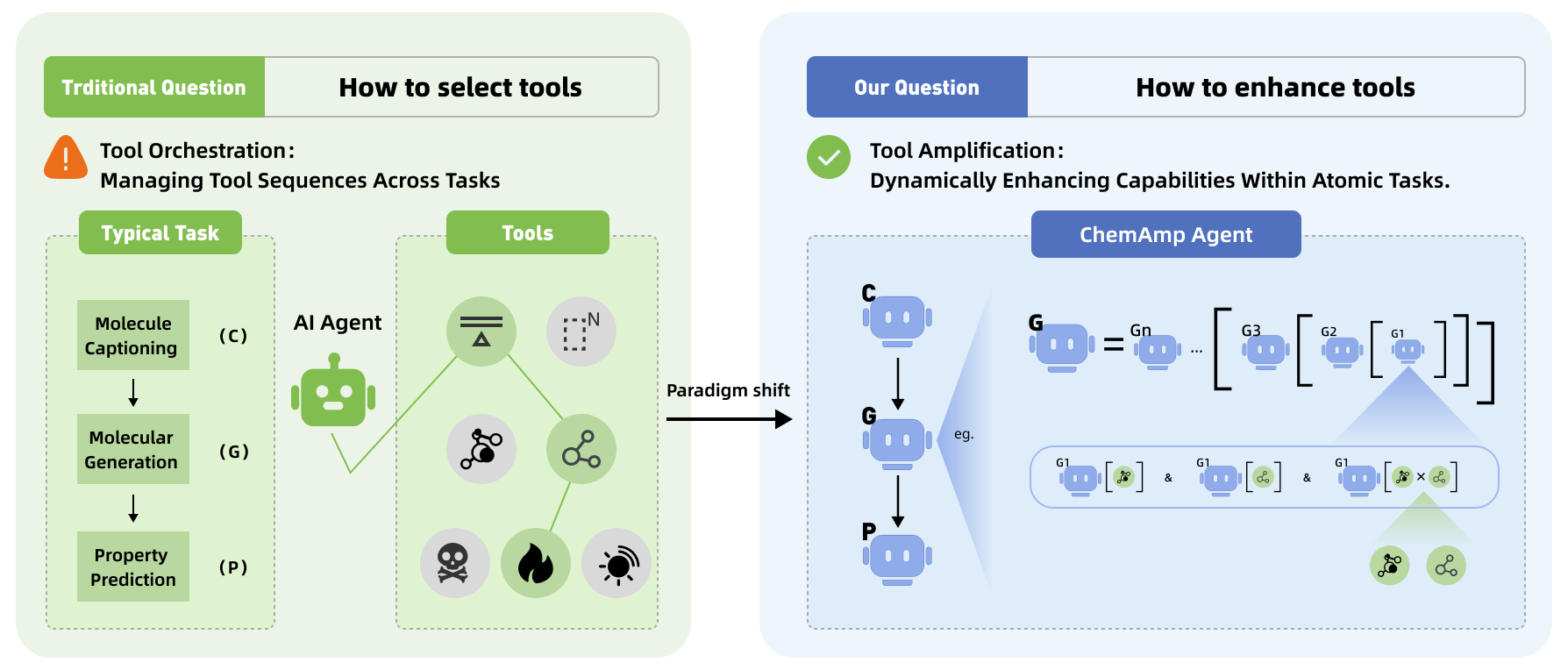}
    \caption{
    The research framework. This work introduces a paradigm shift—from tool orchestration (managing tool sequences across tasks) to tool amplification (dynamically enhancing capabilities within atomic tasks). }
    \label{fig:intro}
\end{figure*}

%% file: figures/method_fig.tex
\begin{figure*}[!ht] 
    \centering
    \includegraphics[width=0.97\textwidth]{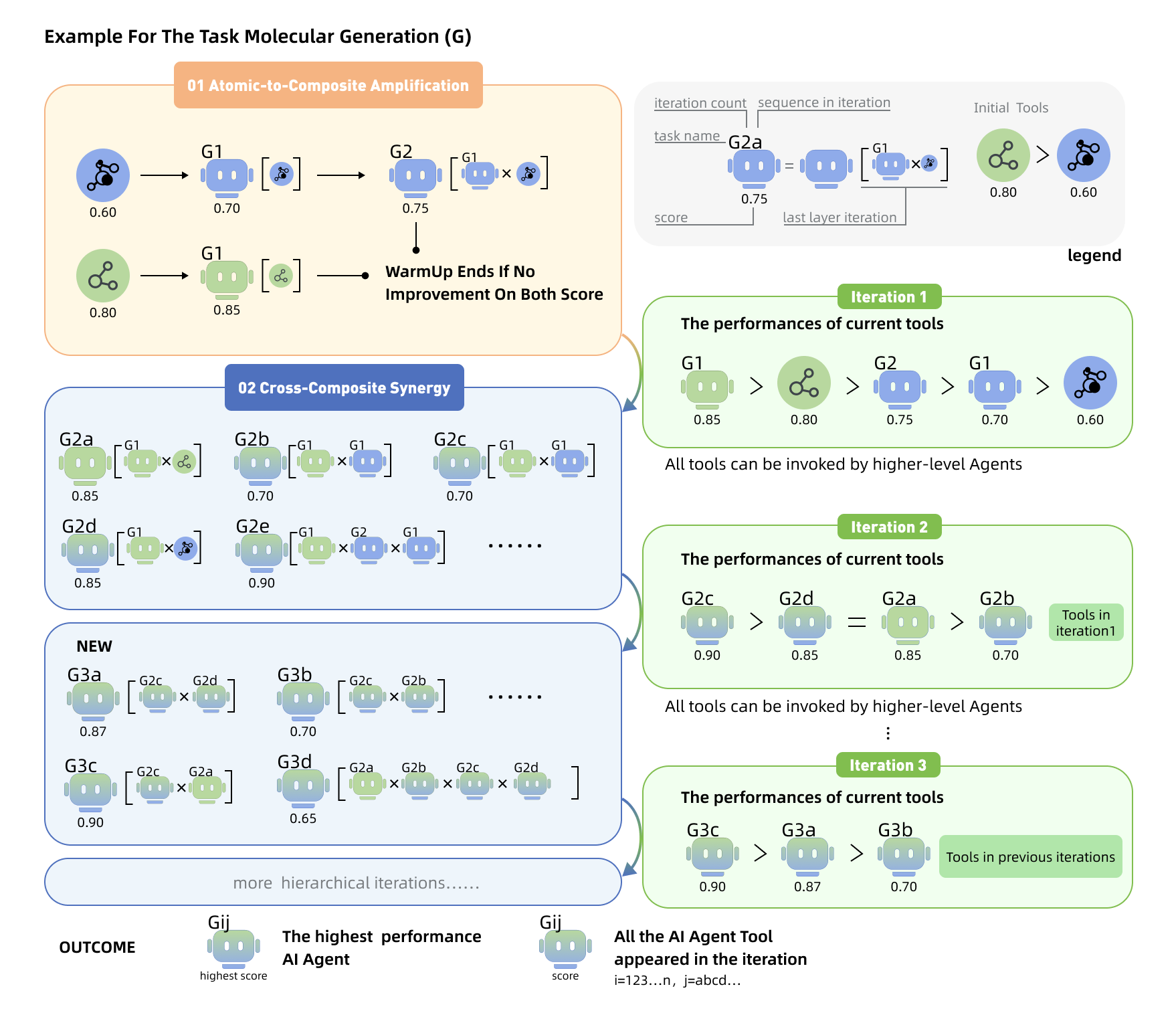}
    \caption{
    ChemAmp's two-stage amplification process:
    (1) Atomic-to-Composite Amplification: atomic tools are encapsulated into agents;
    (2) Cross-Composite Synergy: agents are merged into hierarchical networks.
    }
    \label{fig:method}
\end{figure*}

%% file: 02related.tex
\subsection{Orchestration Paradigm}

The foundation for computational chemistry agents builds on the ReAct framework ~\citep{yao2023reactsynergizingreasoningacting}, which integrates reasoning with tool execution. This paradigm has evolved into sophisticated tool orchestration systems:
(1) Workflow-focused: ChemCrow ~\citep{bran2023chemcrowaugmentinglargelanguagemodels} and Coscientist ~\citep{boiko2023autonomous} sequence tools like RDKit/LabX across multi-step tasks.
(2) Domain-optimized: ChemAgent ~\citep{tang2025chemagentselfupdatinglibrarylarge} and SciToolAgent ~\citep{chen2025scitoolagent} improves LLM performance in complex chemical reasoning tasks by introducing a dynamic, self-evolving memory library that supports task decomposition and solution generation. ChemToolAgent ~\citep{yu2025chemtoolagentimpacttoolslanguage} supports a large tool set and performs dynamic tool selection in a broad task suite.
(3) Hybrid optimization: Recent systems like AgentPrune ~\citep{zhang2024cut}, GPTSwarm ~\citep{zhuge2024gptswarm}, Aflow ~\citep{zhang2024aflow}, and MaAS ~\citep{zhang2025multi} automate workflow refinement after-initial manual design. As a generic optimization of manually orchestrated vanilla multi-agent systems, such frameworks deliberately avoid atomic tool-level performance enhancements. Consequently, our experiments compare against multiple instantiations of vanilla multi-agent systems rather than these derivative approaches.

\subsection{Paradigm Shift Imperative}
Despite these advances, orchestration faces fundamental constraints:
(1) Capability Ceiling: Exact accuracy plateaus at 35\% (e.g., molecule captioning)
(2) Coordination Scope: They remain inefficient in utilizing existing computational chemistry tools~\citep{shi2023relmleveraginglanguagemodels}, and struggle to navigate the combinatorial and hierarchical relationships between these tools.

This impasse necessitates a paradigm shift: from scheduling tools across tasks (orchestration) to amplifying capabilities within tasks through dynamic agent composition. 
The amplification paradigm introduces two core challenges:
(1) Structural Search Complexity: Optimal agent-tool compositions vary significantly across chemistry tasks due to domain-specific constraints.
(2) Efficiency-Capability Tradeoff: Manual composition is infeasible, while exhaustive search incurs prohibitive computational costs.

%% file: 04method.tex
Our study proposes \textbf{ChemAmp}, a framework that realizes the tool amplification paradigm through hierarchical composition of composable agents. ChemAmp dynamically construct agent hierarchies, where atomic tools are iteratively encapsulated into sub-agents and further merged into composite networks. This two-stage process (Fig. \ref{fig:method}) amplifies capabilities via emergent coordination while minimizing task-specific errors through iterative refinement.  

Central to this approach is the \textbf{Agent Composite Tool} $\mathcal{A}(t_1,...,t_n)$, which serves dual roles: as a composable building block for higher-level agents and as an autonomous executor of chemistry sub-tasks. This duality enables ChemAmp to identify optimal capability enhancement points where tool coordination generates synergistic effects beyond individual functions.

\subsection{Stage 1: Atomic-to-Composite Amplification}

To reduce prediction errors of base tools and enrich the Tool Set for Stage 2—thereby constructing a more effective search space for hierarchical coordination—we initiate a process as the warm-up phase. Given an initial tool set $\mathcal{T}$ provided by a large language model and parameters $k$, we construct a global Tool Library $\mathcal{L}$ through Algorithm~\ref{alg:two-stage}. This process begins by applying Atomic-to-Composite Amplification to each atomic tool $t_k \in \mathcal{T}$, where we iteratively construct \textbf{Agent Composite Tools} $\mathcal{A}_i(t_k)$ through layered encapsulation. Each composite tool's performance is evaluated via task-specific metrics to obtain capability enhancement scores $s_i$, with progression to the next reinforcement layer $\mathcal{A}_{i+1}(t_k,\mathcal{A}_i)$ contingent on $s_i$ surpassing the previous layer's score $s_{i-1}$ by a significance threshold $\delta$.

The iterative refinement continues until performance plateaus—defined as $\Delta s < \delta$ for consecutive iterations—ensuring optimal capability enhancement is achieved. Upon termination, all reinforced Agent Composite Tools $\mathcal{A}_{i+1}(t_k,\mathcal{A}_i)$ alongside original base tools are registered in $\mathcal{L}$, creating an enriched library containing context-adapted tool variants, multi-layer composites with emergent functionalities, and error-corrected versions minimizing prediction inaccuracies. This output $\mathcal{L}$ serves as the fundamental building block repository for Stage 2's hierarchical coordination, enabling the construction of task-optimized tool networks.

\input{tables/algo}
\subsection{Stage 2: Cross-Composite Synergy}
After the Stage 1, we obtain a richer tool library $\mathcal{L}$ for performing hierarchical reinforcement through stacking. In this stage, the primary focus is on combining and stacking tools from the library $\mathcal{L}$ to further enhance their performance. First, sort the tool library $\mathcal{L}$, and the top performing tool (top 1) is selected as the mandatory base tool $t_1$. It is then combined with the remaining top-k tools $t_{topk}=\{t_2,t_3,..,t_k\}$ to form a set of agent tools $\{\mathcal{A}(t_1,t_2),...,\mathcal{A}(t_1,t_k)\}$. The best score $s_i$ from this set is compared with the global highest score $s^*$. If the new score exceeds the current highest, another layer of stacking is performed which is also added to the library $\mathcal{L}$. This process is repeated iteratively. Ultimately, the most effective agent composite tool is selected for experimentation.

%% file: tables/algo.tex
\begin{algorithm}[H]
\caption{ChemAmp}
\label{alg:two-stage}
\begin{algorithmic}[1]
\REQUIRE Initial Tool Set $\mathcal{T}$, Param $k$, Tool Library $\mathcal{L}\gets\emptyset$
\ENSURE Best Agent Composite Tool$\mathcal{A}^*$

\STATE \textbf{[Stage 1: Atomic-to-Composite Amplification]}
\FOR{each $t\in\mathcal{T}$}               
  \REPEAT                                 
    \STATE Build and validate agent composite tool
    $\mathcal{A}_n(t,\mathcal{A}_{n-1}(t))$
  \UNTIL{no score improvement}
  \STATE $\mathcal{L}\gets\mathcal{L}\cup\{t,\mathcal{A}_1,...,\mathcal{A}_n\}$
\ENDFOR

\STATE \textbf{[Stage 2: Cross-Composite Synergy]}
\WHILE{global performance improvement}
  \STATE Sort $\mathcal{L}$ by performance; pick {$t_1$} and top-$k$; 
  \STATE Build and validate $\{\mathcal{A}_1(t_1,t_2),...,\mathcal{A}_k(t_1,t_k) \}$
  \STATE $\mathcal{L}\gets\mathcal{L}\cup \{t_1,...,t_k,\mathcal{A}_1,...,\mathcal{A}_k \}$
\ENDWHILE
\end{algorithmic}
\end{algorithm}

%% file: 05experiment.tex
\subsection{Experiment Setup}

\paragraph{Dataset}

We evaluate the performance of ChemAmp in the field of chemistry using ChemLLMBench. ChemLLMBench~\citep{guo2023largelanguagemodelschemistry} comprises a series of chemistry-related tasks that cover a wide range of chemical topics. 
In this study, we focus on four representative tasks and select 100 evaluation instances for each task, consistent with the evaluation experiments in ChemDFM, as the test set.
Since the evaluation dataset for each task in ChemLLMBench contains only 100 instances, we adopt a similar approach to that in ~\citep{guo2023largelanguagemodelschemistry} to select the validation set. 
For the Text-Based Molecule Design and Molecule Captioning tasks, we randomly sample 100 instances from the ChEBI-20-MM~\citep{liu2025quantitativeanalysisknowledgelearningpreferences}
dataset, excluding the corresponding test set, as the sample set for validation. 
For the Reaction Prediction task, we randomly sample 100 instances from the USPTO-MIT~\citep{jin2017predictingorganicreactionoutcomes} dataset, excluding the corresponding test set, as the validation set. 
For the Molecular Property Prediction task, we randomly sample 50 instances for each dataset from the BBBP, HIV, BACE, Tox21, and ClinTox~\citep{wu2018moleculenetbenchmarkmolecularmachine} datasets, excluding the corresponding test sets, as the validation set.
The details of our dataset are shown in supplementary material.

\paragraph{Models}
We categorize current models into two primary groups: chemistry specialized models and generalist LLMs. chemistry specialized models refer to non-LLM models designed for specific tasks, while generalist LLMs are LLMs primarily designed for general-purpose use across a wide range of tasks. And the agent in the LLM-based model leverage GPT-4o~\citep{openai2024gpt4ocard} as the core agent with ReAct~\citep{yao2023reactsynergizingreasoningacting} framework, augmented with additional tools tailored to the specific task. To ensure fair comparisons, we use the same test set for evaluating different models on each task.
\paragraph{Tools}

For each task, an initial tool set is provided. The large language model autonomously selects the most suitable tool from the tool set as the base tool to be enhanced. In the following experiments, we report only the best-performing results.

\subsection{Results}
\subsubsection{Text-based Molecule Design}

In the text-based molecule design task, LLMs predict a molecule’s SMILES (Simplified Molecular Input Line Entry System) representation based on a given description, testing their ability to interpret and translate chemical language into valid molecular structures~\citep{zhao2024chemdfmlargelanguagefoundation}.

Our study employs two sets of metrics to evaluate the performance of the task. The first set of metrics measures the text-based similarity between the predicted SMILES and the gold standard SMILES, including exact match, BLEU, and Levenshtein distance~\citep{haldar2011levenshteindistancetechniquedictionary}. 
The second set of metrics assesses the chemical similarity between the predicted molecules and the reference molecules, encompassing the validity of the predicted SMILES and the FTS (Fingerprint Tanimoto Similarity)~\citep{tanimoto1958elementary}, calculated based on MACCS, RDK, and Morgan~\citep{Morgan1965TheGO}.

\input{tables/text}

From the results in Tab.~\ref{tab:text_based}, our method consistently outperforms all baselines. It achieves high molecular validity while satisfying target specifications. In contrast, LLM-based models exhibit limited performance, particularly in Exact and BLEU scores, underscoring the difficulty of accurate molecular generation in a 0-shot setting. 

\subsubsection{Molecule Captioning}

To evaluate the model’s capacity to translate complex chemical representations into natural language, we adopt the Molecule Captioning task~\citep{guo2023largelanguagemodelschemistry}, which requires generating concise descriptions from input SMILES strings. 
To assess the model's performance on this task, we employ traditional NLP evaluation metrics, such as BLEU and ROUGE, to measure the similarity between the molecule descriptions generated by the model and the reference descriptions in the test set.
\input{tables/caption}

As shown in Tab.~\ref{tab:captioning}, ChemAmp framework, achieves state-of-the-art performance across all metrics. Chemistry specific models maintain a clear advantage in molecule captioning, while most general-purpose LLMs—except large-scale ones like GPT-4o and DeepSeek-R1—exhibit results. 

\subsubsection{Molecular Property Prediction}
Molecular property prediction~\citep{Guo_2021,wang2021chemicalreactionawaremoleculerepresentationlearning} is a core task in computational chemistry, with broad applications in drug discovery and materials science. This task involves predicting chemical or physical properties from molecular structures. We evaluate performance on five benchmarks from MoleculeNet~\citep{wu2018moleculenetbenchmarkmolecularmachine}: BACE, BBBP, HIV, ClinTox, and Tox21. Considering that our method is based on large language models, we adopt accuracy as the primary evaluation metric.

\input{tables/prediction}

\input{figures/Validationcost}
As shown in Tab.~\ref{tab:prediction}, ChemToolAgent achieved the best performance on the BBBP and Clintox datasets, which benefits from the capabilities of the tools it employs. Conversely, our method achieved state-of-the-art results on the other three datasets.
Utilizing only the two open-source models, Uni-Mol-v2 and ChemDFM-13B, our approach demonstrates evident improvements over both models individually.

\subsubsection{Reaction Prediction}
Reaction prediction is a fundamental task in chemistry, essential for drug discovery, materials science, and the design of novel synthetic routes. Given a set of reactants, the objective is to predict the most probable reaction products~\citep{guo2024modeling,Schwaller_2019}. Consistent with the text-based molecule design task, we employ the same evaluation metrics to assess model performance.
\input{tables/reaction}

As shown in Tab.~\ref{tab:reaction_prediction}, Chemformer demonstrates strong performance on this task, achieving a product prediction accuracy of 0.91 and outperforming other task-specific models across all metrics. In contrast, LLMs face notable challenges; for instance, DeepSeek-R1 achieves only 0.10 accuracy despite its advanced reasoning capabilities, and ChemDFM performs poorly under zero-shot settings. ChemAmp achieves competitive results across all metrics, surpassing Chemformer in all but Exact score (with only a marginal 0.01 difference).

%% file: tables/text.tex
\setlength\tabcolsep{1pt}
\begin{table}[!htb]
    \centering
    \small
    \definecolor{lightgray}{gray}{0.9} 
    \resizebox{0.48\textwidth}{!}{

    \begin{tabular}{lcccccccc}
        \toprule
        \textbf{Model} & \textbf{Exact} & \textbf{BLEU} & \textbf{Dis} & \textbf{Validity} & \textbf{MACCS} & \textbf{RDK} & \textbf{Morgan} & \textbf{FTS} \\
        \midrule
        \rowcolor{lightgray} \multicolumn{9}{c}{\textit{Chemistry specialized models}} \\

        MolXPT & 0.22 & - & - & \underline{0.98} & 0.86 & 0.76 & 0.67 & 0.76 \\
        Text+Chem T5 & 0.32 & 0.85 & 16.87 & 0.94 & \underline{0.90} & \underline{0.82} & \underline{0.75} & 0.82 \\
        Mol-Instruction& 0.02 & 0.35 & 41.40 & \textbf{1.00} & 0.41 & 0.23 & 0.15 & 0.26 \\
        ChemDFM-13B& 0.32 & 0.85 & \underline{11.58} & 0.94 & 0.81 & 0.73 & 0.67 & 0.74 \\
        \midrule
        \rowcolor{lightgray} \multicolumn{9}{c}{\textit{Generalist LLMs}} \\
        GPT-4o & 0.01 & 0.57 & 52.85 & 0.91 & 0.71 & 0.54 & 0.38 & 0.54 \\
        Deepseek-R1 & 0.02 & 0.56 & 92.29 & 0.57 & 0.48 & 0.38 & 0.31 & 0.39 \\
        Llama3-70b & 0.03 & 0.57 & 46.63 & 0.78 & 0.57 & 0.40 & 0.30 & 0.42 \\
        \midrule
        \rowcolor{lightgray} \multicolumn{9}{c}
        {\textit{Agent with Tool Orchestration}} \\
        ChemCrow & 0.34 & 0.77 & 45.4 & 0.68& 0.84 & 0.80 & 0.76 & 0.80 \\
        ChemToolAgent & 0.28 & - & - & 0.90 &- & - &- & 0.77 \\
        \midrule
        \rowcolor{lightgray} \multicolumn{9}{c}
        {\textit{Agent with Tool Amplification}} \\
        Ours (Stage 1) & \underline{0.34} & \underline{0.87} & 12.63 & 0.94 & 0.85 & 0.80 & 0.74 & 0.80 \\
        Ours (Stage 1+2) & \textbf{0.38} & \textbf{0.93} & \textbf{8.68} & 0.96 & \textbf{0.92} & \textbf{0.87} & \textbf{0.80} & \textbf{0.86} \\

        \bottomrule
    \end{tabular}
    }
    
    \caption{Benchmark results in text-based molecule design task. All methods are evaluated on 0-shot. Our approach utilizes the chemistry-specialized models listed in this table as tools. ChemCrow and ChemToolAgent additionally employ all tools mentioned in their papers, amounting to 18 and 29 tools respectively.}
    \label{tab:text_based}
\end{table}

%% file: tables/caption.tex
\setlength\tabcolsep{1pt}
\begin{table}[!htb]
    \centering
    \small
    \definecolor{lightgray}{gray}{0.9} 
    \resizebox{0.47\textwidth}{!}{
    \begin{tabular}{lccccc}
        \toprule
        \textbf{Model} & \textbf{BLEU-2} & \textbf{BLEU-4} & \textbf{ROUGE-1} & \textbf{ROUGE-2} & \textbf{ROUGE-L} \\
        \midrule
        \rowcolor{lightgray}\multicolumn{6}{c}{\textit{Chemistry specialized models}} \\
        Text+Chem T5 & 0.63 & 0.54 & \underline{0.68} & \underline{0.54} & \underline{0.62} \\
        MolXPT & 0.59 & 0.50 & 0.66 & 0.51 & 0.60 \\
        InstructMol & 0.48 & 0.37 & 0.57 & 0.39 & 0.50 \\
        Mol-Instruction & 0.25 & 0.17 & 0.33 & 0.29 & 0.27 \\
        ChemDFM-13b & 0.32 & 0.27 & 0.49 & 0.37 & 0.48 \\
        \midrule
        \rowcolor{lightgray}\multicolumn{6}{c}{\textit{Generalist LLMs}} \\
        GPT-4o & 0.26 & 0.17 & 0.10 & 0.00 & 0.30 \\
        Deepseek-R1 & 0.40 & 0.25 & 0.10 & 0.02 & 0.21 \\
        Llama3-70b & 0.11 & 0.07 & 0.06 & 0.00 & 0.12 \\
        \midrule
        \rowcolor{lightgray}\multicolumn{6}{c}{\textit{Agent with Tool Orchestration}} \\
        ChemCrow & 0.25 & 0.20 & 0.18 & 0.11 & 0.23 \\
        ChemToolAgent &0.63 & 0.54&0.45&0.31&0.55\\
        \midrule
        \rowcolor{lightgray}\multicolumn{6}{c}{\textit{Agent with Tool Amplification}} \\
        Ours (Stage 1) & \underline{0.64} & \underline{0.56} & 0.45 & 0.29 & 0.55 \\
        Ours (Stage 1+2) & \textbf{0.73} & \textbf{0.69} & \textbf{0.70} & \textbf{0.58} & \textbf{0.76} \\
        \bottomrule
        \end{tabular}
    }
    \caption{Benchmark results in molecule captioning task. All methods are evaluated on 0-shot. Our approach utilizes the chemistry-specialized models listed in this table as tools. ChemCrow and ChemToolAgent additionally employ all tools mentioned in their papers, amounting to 18 and 29 tools respectively.}
    \label{tab:captioning}
\end{table}

%% file: tables/prediction.tex
\setlength\tabcolsep{1pt}
\begin{table}[!htb]
    \centering
    \small
    \definecolor{lightgray}{gray}{0.9} 
    \resizebox{0.45\textwidth}{!}{
    \begin{tabular}{lcccccc}
        \toprule
        \textbf{Model} & \textbf{BACE} & \textbf{BBBP} & \textbf{Clintox} & \textbf{HIV} & \textbf{Tox21} & \textbf{Avg}\\
        \midrule
        \rowcolor{lightgray}\multicolumn{7}{c}{\textit{Chemistry specialized models}} \\
Uni-Mol-v2	&0.75&	0.58&	\underline{0.51}	&\underline{0.96}	&0.92&	\underline{0.74} \\
ChemDFM-13B & 0.66 & 0.57 & 0.49    & 0.94 & 0.83  & 0.70\\
        \midrule
        \rowcolor{lightgray}\multicolumn{7}{c}{\textit{Generalist LLMs}} \\
GPT-4o    & 0.38 & 0.56 & 0.51    & 0.59 & 0.37  & 0.48 \\
Deepseek-R1 & 0.62 & 0.61 & 0.48    & 0.51 & 0.75  & 0.60 \\
Llama3-70B  & 0.55 & 0.59 & 0.48    & 0.20 & 0.59  & 0.48 \\
        \midrule
        \rowcolor{lightgray}\multicolumn{7}{c}{\textit{Agent with Tool Orchestration}} \\
ChemCrow  & 0.61 & 0.60 & 0.50    & 0.60 & 0.77  & 0.62 \\
ChemToolAgent & - & \textbf{0.90} & \textbf{0.82} & 0.94 & - & - \\
        \midrule
        \rowcolor{lightgray}\multicolumn{7}{c}{\textit{Agent with Tool Amplification}} \\
Ours (Stage 1)                  & \underline{0.75} & \underline{0.59} & 0.49    & 0.92 & \underline{0.94}  & 0.74 \\
Ours (Stage 1+2)           & \textbf{0.79} & 0.68 & 0.67    & \textbf{0.96} & \textbf{0.96}  & \textbf{0.81}\\

        \bottomrule
        \end{tabular}
    }
    \caption{Benchmark results in molecular property prediction task. All methods are evaluated on 0-shot. Our approach utilizes the chemistry-specialized models listed in this table as tools. ChemCrow and ChemToolAgent additionally employ all tools mentioned in their papers, amounting to 18 and 29 tools respectively.}
    \label{tab:prediction}
\end{table}

%% file: figures/Validationcost.tex
\begin{figure*}[t] 
    \centering
    \includegraphics[width=1\textwidth]{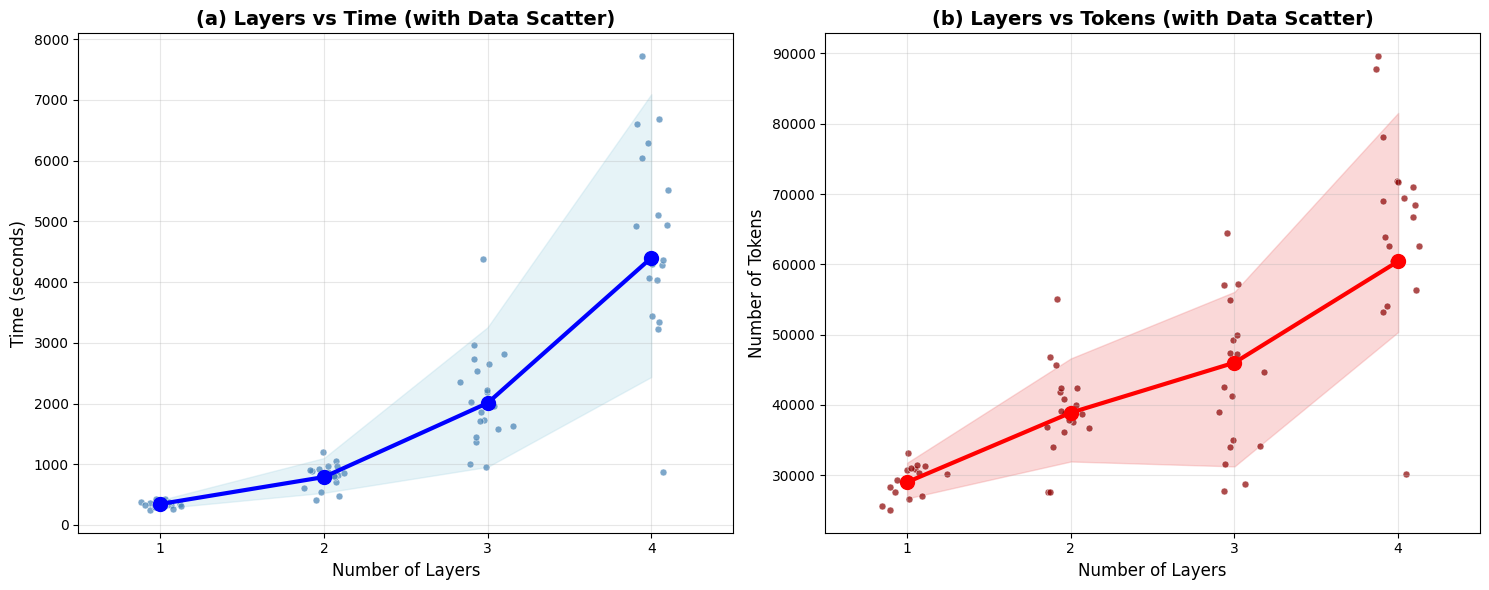}
    \caption{
    Validation cost (time, tokens) over all tasks with increasing layer depth.
    }
    \label{fig:Validation cost}
\end{figure*}

%% file: tables/reaction.tex
\setlength\tabcolsep{1pt}
\begin{table}[!htb]
    \centering
    \small
    \definecolor{lightgray}{gray}{0.9} 
    \resizebox{0.47\textwidth}{!}{
    \begin{tabular}{lcccccccc}
        \toprule
        \textbf{Model} & \textbf{Exact} & \textbf{BLEU} & \textbf{Dis} & \textbf{Validity} & \textbf{MACCS} & \textbf{RDK} & \textbf{Morgan} & \textbf{FTS} \\
        \midrule
        \rowcolor{lightgray} \multicolumn{9}{c}{\textit{Chemistry specialized models}} \\

        Chemformer & \underline{0.91} & 96.1 & \underline{1.26} & 1.00 & \underline{0.97} & \underline{0.97} & \underline{0.96} & 0.97 \\
        Text+ChemT5 & 0.83 & 96.0 & 7.42 & 0.98 & 0.96 & 0.96 & 0.94 & 0.95 \\
        InstructMol & 0.54 & 96.7 & 10.85 & 1.00 & 0.88 & 0.78 & 0.74 & 0.80 \\
        Mol-Instruction & 0.05 & 65.4 & 27.26 & 1.00 & 0.51 & 0.31 & 0.26 & 0.36 \\
        ChemDFM-13B & 0.39 & 80.6 & 10.38 & 0.96 & 0.77 & 0.69 & 0.65 & 0.70 \\
        \midrule
        \rowcolor{lightgray} \multicolumn{9}{c}{\textit{Generalist LLMs}} \\
        GPT-4o & 0.01 & 65.8 & 27.24 & 0.81 & 0.54 & 0.39 & 0.33 & 0.42 \\
        Deepseek-R1 & 0.10 & 76.2 & 16.04 & 0.75 & 0.60 & 0.53 & 0.48 & 0.54 \\
        Llama3-70b & 0.00 & 55.2 & 282.46 & 0.85 & 0.48 & 0.35 & 0.31 & 0.38 \\
        \midrule
        \rowcolor{lightgray} \multicolumn{9}{c}{\textit{Agent with Tool Orchestration}} \\
        ChemCrow & 0.70 & 90.2 & 10.11 & 0.90 & 0.95 & 0.93 & 0.90 & 0.93 \\
        ChemToolAgent & 0.78 & - & - & 0.98 & - & - & - & 0.98 \\
        \midrule
        \rowcolor{lightgray} \multicolumn{9}{c}{\textit{Agent with Tool Amplification}} \\
        Ours (Stage 1) & 0.87 & \underline{97.1} & 1.00 & 1.00 & 0.97 & 0.97 & 0.95 & 0.96 \\
        Ours (Stage 1+2) & \textbf{0.91} & \textbf{98.4} & \textbf{0.97} & \textbf{1.00} & \textbf{0.98} & \textbf{0.98} & \textbf{0.96} & \textbf{0.98} \\

        \bottomrule
    \end{tabular}
    }
    \caption{Benchmark results in reaction prediction task. All methods are evaluated on 0-shot. Our approach utilizes the chemistry-specialized models listed in this table as tools. ChemCrow and ChemToolAgent additionally employ all tools mentioned in their papers, amounting to 18 and 29 tools respectively.}
    \label{tab:reaction_prediction}
\end{table}

%% file: ana.tex
\subsection{Generalization: Few-Shot\&Tool-Constrained}

To evaluate the influence of tool diversity and validation data size, we conduct experiments on the text-based molecule design task. As shown in Tab.~\ref{tab:analysis1}, increasing the number of tools from 2 to 4 under a fixed validation set leads to minimal variation in BLEU-2 scores (0.86 for Tool Number = 2 and 4; 0.85 for Tool Number = 3), suggesting diminishing returns from tool addition.

Additionally, we examine the effect of validation data size by comparing BLEU-2 scores under varying data volumes. Scores generally increase with more data—for instance, with Tool Number = 2, BLEU-2 improves from 0.79 (data = 5) to 0.89 (data = 10).

\input{tables/ana1}

\input{tables/cost}

\input{tables/ana2}

\subsection{Validation cost over the four tasks}
We validated the composite structure bottom-up, akin to a greedy search. The time and token cost during validation was measured across four tasks. As illustrated in Fig.~\ref{fig:Validation cost}, the cost exhibits near-linear growth within the first three layers (Performance stabilizes post-convergence in our tasks), transitioning to a steep increase beyond the fourth layer.

\subsection{Layer Depth - Performance Scaling}
This section investigates the impact of layer depth on performance in the text-based molecule design task. As shown in Tab.~\ref{tab:analysis2}, increasing the number of stacking layers initially improves BLEU-2 scores, with performance peaking at seven layers. Shallower architectures (e.g., 1–2 layers) may lack the capacity to model complex feature interactions, while deeper configurations beyond seven layers yield diminishing or even negative returns. This suggests a performance plateau likely caused by redundant information or overfitting in overly deep structures.

\subsection{Why does the agent composite tool work?}
\label{6.4}

From the case studies, we identify four distinct behavior patterns—correct, modify, judge and reserve, with detailed information on other cases provided in supplementary material. 
These findings demonstrate the agent-composite tool's capability to aggregate information from diverse tool structures while simultaneously filtering erroneous outputs.
In the text-based molecular design task, initially, Tool A predicts the SMILES string ``C1(C)CC1'' for cyclopropane, which is incorrect due to a misinterpretation of the three-membered ring, introducing an erroneous branch. To correct this, Tool B is invoked with the query ``Cyclopropane'' and returns the correct SMILES ``C1CC1'', consistent with the standard structure.

\subsection{Inference Cost vs. Multi-Agent Networks}
Comparison of Multi-Agent Systems and Tool-Augmented LLMs
Multi-agent systems (MAS) based on LLMs and our method share some characteristics, including task decomposition, tool invocation, and information sharing. Our study evaluates the performance of six baseline MAS with varied communication architectures against our method in a text-based molecular design task. In Figure~\ref{tab:multiagent}, our method achieves substantial efficiency improvements over baseline multi-agent networks. In a four-agent configuration, it reduces token consumption to one-eighteenth of the average for MAS while preserving comparable inference latency. This efficiency stems from our method's optimized coordination of specialized tools, enabling robust performance with minimal computational overhead. Further exploration of these coordination mechanisms could enhance their applicability to other complex scientific tasks. Detailed descriptions of the MAS configurations are available in the Appendix.

%% file: tables/ana1.tex
\setlength\tabcolsep{1pt}
\begin{table}[!htb]
    \centering
    \small
    \definecolor{lightgray}{gray}{0.9} 
    \resizebox{0.48\textwidth}{!}{
    \begin{tabular}{ccccccccccc}
        \toprule
        \multirow{2}{*}{\textbf{Tool Number}} & \multicolumn{8}{c}{\textbf{Sample Size of Validation Data}} & \multirow{2}{*}{\textbf{AVG}} \\ 
        \cmidrule(lr){2-9}
        & \multicolumn{2}{c}{5} & \multicolumn{2}{c}{10} & \multicolumn{2}{c}{20} & \multicolumn{2}{c}{30} & \\ 
        \cmidrule(lr){2-3} \cmidrule(lr){4-5} \cmidrule(lr){6-7} \cmidrule(lr){8-9}
        & Layer & BLEU-2 & Layer & BLEU-2 & Layer & BLEU-2 & Layer & BLEU-2 & \\
        \midrule
        2 & 0.6 & 0.79 & 2.7 & 0.89 & 3.0 & 0.89 & 3.2 & 0.87 & 0.86 \\
        3 & 0.4 & 0.81 & 2.5 & 0.87 & 2.8 & 0.86 & 3.0 & 0.85 & 0.85 \\
        4 & 0.6 & 0.80 & 2.8 & 0.88 & 3.0 & 0.87 & 3.2 & 0.86 & 0.86 \\
        \bottomrule
    \end{tabular}
    }
    \caption{Comparison of the performance of different tools on the text-based molecule design task under different validation data sizes. ``Layer'' represents the average number of layers of agent composite Tools.}
    \label{tab:analysis1}
\end{table}

%% file: tables/cost.tex
\setlength\tabcolsep{1pt}
\begin{table*}[!htb]
    \centering
    \tiny
    \definecolor{lightgray}{gray}{0.9} 
    \resizebox{0.80\textwidth}{!}{
\begin{tabular}{cccccccccc}
\toprule
\rowcolor{lightgray} \multicolumn{10}{c}{Text-based Molecule Design (BLEU-2) - w/o-Tool} \\ 
\midrule
NUM  & Chain  & Random  & FullConnected  & Layered  & Star   & Debate  & Avg   & Avg\_all\_tokens & Avg\_Time \\ 
\midrule
0    & 0.548  & 0.548   & 0.548          & 0.548    & 0.548  & 0.548   & 0.548 & 890.19           & 8.254     \\
1    & 0.604  & 0.604   & 0.604          & 0.604    & 0.604  & 0.604   & 0.604 & 2315.22          & 15.826    \\
2    & 0.587  & 0.587   & 0.587          & 0.587    & 0.587  & 0.587   & 0.587 & 3619.54          & 21.034    \\
4    & 0.608  & 0.642   & 0.672          & 0.666    & 0.661  & 0.639   & 0.648 & 52010.75         & 74.029    \\
8    & 0.633  & 0.701   & 0.684          & 0.640    & 0.649  & 0.636   & 0.657 & 344604.597       & 227.012   \\ 
\midrule
\rowcolor{lightgray} \multicolumn{10}{c}{Text-based Molecule Design (BLEU-2) - Tool (Name2SMILES,ChemDFM)} \\ 
\midrule
NUM  & Chain  & Random  & FullConnected  & Layered  & Star   & Debate  & Avg   & Avg\_all\_tokens & Avg\_Time \\ 
\midrule
0    & 0.832  & 0.832   & 0.832          & 0.832    & 0.832  & 0.832   & 0.832 & 2528.79          & 21.382    \\
1    & 0.842  & 0.842   & 0.842          & 0.842    & 0.842  & 0.842   & 0.842 & 3125.64          & 30.237    \\
2    & 0.844  & 0.844   & 0.844          & 0.844    & 0.844  & 0.844   & 0.844 & 18446.91         & 108.427   \\
4    & 0.864  & 0.888   & 0.885          & 0.867    & 0.884  & 0.889   & 0.880 & 150544.061       & 2207.641  \\
8    & 0.841  & 0.890   & 0.854          & 0.887    & 0.891  & 0.874   & 0.873 & 733540.998       & 6744.788  \\ 
\midrule

NUM  & \multicolumn{6}{c}{Our Agent Composite Tool}                                                                 & Score & Avg\_all\_tokens & Avg\_Time \\ 
\midrule
2    & \multicolumn{6}{c}{{[}'ChemDFM\_2'{]}}                                                            & 0.898 & 2801.93          & 39.103    \\
3    & \multicolumn{6}{c}{{[}'Name2SMILES\_1','ChemDFM\_1'{]}}                                           & 0.918 & 2821.38          & 63.795    \\
4    & \multicolumn{6}{c}{{[}'Name2SMILES\_3','ChemDFM\_0'{]}}                                           & 0.904 & 2745.07          & 78.144    \\
4    & \multicolumn{6}{c}{{[}{[}'ChemDFM\_0', 'Name2SMILES\_1'{]}, 'ChemDFM\_1'{]}}                      & \textbf{0.925} & 2851.43          & 72.484    \\
8    & \multicolumn{6}{c}{\begin{tabular}[c]{@{}c@{}}{[}{[}'ChemDFM\_1','Name2SMILES\_1'{]},\\ {[}'ChemDFM\_1','Name2SMILES\_2'{]}{]}\end{tabular}} & 0.907 & 1830.11          & 95.623   \\
\bottomrule
\end{tabular}
    }
    \caption{Inference Cost vs. Multi-Agent Networks. Each tool's annotated depth indicates its layer depth after Stage 1.  Name2SMILES is a basic retrieve tool for chemistry online database widely used in chemistry agent like ChemCrow.}
    \label{tab:multiagent}
\end{table*}

%% file: tables/ana2.tex
\setlength\tabcolsep{1pt}
\begin{table}[!htb]
    \centering
    \small
    \definecolor{lightgray}{gray}{0.9} 
    \begin{tabular}{lccccccccc}
    \toprule
    Layer & 1 & 2 & 3 & 4 & 7 & 8 & 10 & 10+ & AVG \\
    \midrule
    BLEU-2 & 0.88 & 0.90 & 0.92 & 0.93 & 0.90 & 0.92 & 0.91 & 0.90 & 0.91 \\
    \bottomrule
    \end{tabular}
    \caption{Comparison of different agent stacking levels on the performance of text-based molecule design tasks.}
    \label{tab:analysis2}
\end{table}

%% file: 06conclusion.tex
We introduced tool amplification for chemistry. It is a novel paradigm that transcends tool orchestration by dynamically amplifying atomic tools' capabilities through hierarchical coordination within individual tasks. Instantiating this paradigm, ChemAmp leverages bi-phase iterative encapsulation to construct task-specialized composable super-agents from atomic tools (e.g., UniMol2, Chemformer). Evaluations across molecular design, captioning, reaction prediction, and property prediction demonstrate state-of-the-art results, outperforming chemistry specialized models, generalist LLMs, and orchestration-based agents. Critically, ChemAmp achieves these gains with minimal data ($\leq10$ samples) while reducing inference token costs by 94\% versus vanilla multi-agent systems. 
This work establishes tool amplification as a scalable, efficient foundation for next-generation LLM-agent systems in complex scientific domains.

\section*{Limitations}
While ChemAmp demonstrates significant advancements in tool amplification for chemistry tasks, its evaluation has primarily focused on four core tasks: molecular design, molecule captioning, reaction prediction, and property prediction. These tasks, though representative, do not encompass the full spectrum of challenges in chemistry or other scientific domains. Extending tool amplification to additional tasks, such as retrosynthetic planning or multi-step reaction optimization, could reveal further opportunities to refine the bi-phase iterative encapsulation process and enhance its robustness across diverse problem spaces.

Additionally, ChemAmp's reliance on a small number of samples ($\leq10$) for constructing task-specialized super-agents, while efficient, assumes high-quality input data tailored to specific tasks. In scenarios where data is noisy, incomplete, or highly heterogeneous, the framework's performance may require further adaptation, such as integrating advanced data preprocessing or transfer learning techniques. Future work could explore these adaptations to broaden the applicability of tool amplification in real-world scientific workflows with varying data constraints.

\section*{Acknowledgments}
This work was supported by the National Natural Science Foundation of China (Grant No. 82394432, and 92249302), and the Shanghai Municipal Science and Technology Major Project (Grant No. 2023SHZDZX02) .

%% file: Appendix.tex
\section{ChemAmp Agent Details}
\label{Stacking Agent Details}
Now we will present detailed results and scores for each agent in the Tab.~\ref{tab:appendix_molecule_design}, along with the corresponding prompts:

\subsection{Agent}
\subsubsection{Prompt}
Our Agent framework is based on the ReAct method to implement tool  and reasoning processes. 
\subsubsection{Naming Rule}
To facilitate the comprehension of ChemAmp, we propose a systematic hierarchical naming rules. In addition, in order to prevent the tool name from affecting the agent call, we choose to anonymously process the agent tool, that is, \{task name\}\_\{num\}.
\begin{itemize}
    \item \textbf{Atomic-to-Composite Amplification}: Hierarchical proxies are constructed through a recursive generation strategy, with the depth of the hierarchy dynamically extendable via the numerical suffix. For instance, "["Name2SMILES\_0"]" denotes the base tool, while "["Name2SMILES\_1"]" signifies a first-layer tool, referred to as an Agent Tool (which encapsulates both the tool and the Agent into a new tool) and "["ChemDFM\_2"]".

    \item \textbf{Cross-Composite Synergy}: The combination of multiple tools within an Agent is represented in a list format, utilizing depth-first traversal to generate sub-tools, thereby forming the final toolset for the agent. For example, the structure "["Name2SMILES\_1", "ChemDFM\_2"]" represents a flat structure with tools at the same level ([A, B, ...]), while the structure "[['Name2SMILES\_0','ChemDFM\_1'],'Nam
    e2SMILES\_1','ChemDFM\_0']" illustrates a nested structure ([[A, B], C, D]), where tools A and B are first combined before being integrated with tool C and D.
\end{itemize}
\input{tables/app1}

\subsubsection{Dataset Setting}
\input{tables/data}
\subsection{Text-based Molecule Design}
\subsubsection{Task Introduction}
The test set of ChEBI-20-MM is exploited for this task in ChemLLMBench. Models are asked to predict the SMILES of the molecule that fits the given description. Considering the low accuracy of the models, we use BLEU-2 as the training metric and use metrics such as Exact, Dis and others during the testing stage.

\subsection{Text-based Molecule Design}
\subsubsection{Task Introduction}
The test set of ChEBI-20-MM is exploited for this task in ChemLLMBench. Models are asked to predict the SMILES of the molecule that fits the given description. Considering the low accuracy of the models, we use BLEU-2 as the training metric and use metrics such as Exact, Dis and others during the testing stage.
\subsubsection{Prompt}
We use a simpler prompt compared with the prompt introduced in \cite{guo2023largelanguagemodelschemistry}
\begin{tcolorbox}[colback=gray!10, colframe=black, title=Prompt: Text-based Molecule Design]
You are an expert chemist. Given the molecular requirements description, your task is to design a new molecule SMILES:\\
Molecular requirements description::
\end{tcolorbox}
\subsection{Molecule Captioning}
\subsubsection{Task Introduction}
The test set  is the same with the Text-based Molecule Design task. Because this is the mirroring task, which generates a detailed description by giving a SMILES to the models. In this task, we also choose the BLEU-2 as the metric in the training stage. When in the test stage, more metrics, like BLEU and ROUGE, are utilized to Measure the performance of the model.
\subsubsection{Prompt}
We also use a simpler prompt compared with the prompt introduced in \cite{guo2023largelanguagemodelschemistry}
\begin{tcolorbox}[colback=gray!10, colframe=black, title=Prompt: Molecule Captioning]
You are an expert chemist. Given the molecular SMILES, your task is to provide the detailed description((The molecule is ...) of the molecule. \\Please strictly follow the format, no other information can be provided.\\Molecular SMILES:
\end{tcolorbox}

\subsection{Molecular Property Prediction}
\subsubsection{Task Introduction}
The molecular property prediction tasks in ChemLLMBench consist of five tasks from MoleculeNet benchmark~\citep{wu2018moleculenetbenchmarkmolecularmachine}, including BACE, BBBP, HIV, ClinTox, and Tox21. Among these, BACE and BBBP are balanced binary classification tasks, while HIV represents an unbalanced binary classification task. ClinTox consists of 2 unbalanced binary classification tasks, and Tox21 comprises 21 unbalanced binary classification tasks. In this task, we choose the AUC-ROC as the first metric in the training stage. Considering the calculation method of AUC-ROC for large language models, we also introduced Accuracy as a second metric for test stage.
\subsubsection{Prompt}
We use the same prompts introduced in ~\citep{guo2023largelanguagemodelschemistry}

\subsection{Reaction Prediction}
\subsubsection{Task Introduction}
The reaction prediction task asks the model to predict the product of the given reaction. ChemLLMBench utilizes the USPTO-MIT dataset for this task. Since the benchmark metric is Accuracy, we also chose Accuracy as the training metric, and considering that the answer is also SMILES, we adopted the same metric as Molecular Design task for measurement during the testing stage.
\subsubsection{Prompt}
We reformat the prompt provided ~\citep{guo2023largelanguagemodelschemistry}.
\begin{tcolorbox}[colback=gray!10, colframe=black, title=Prompt: Reaction Prediction]
Given an incomplete chemical reaction equation in SMILES notation (format: reactants>>product, where multiple reactants are separated by dots '.'), predict and complete the missing products marked as '\_\_\_'. The response should only contain the only one SMILES representation of the missing molecule, without any additional explanation (Note: Please only output only one final product). Please answer the question based on the following Chemical reaction equation:
\end{tcolorbox}

\section{Multi-agent Implementation Details}

In this section, we will discuss how to implement multi-agent systems and specific ways of information transmission, including chain, random, star, full-connected, layered, and debate graphs.

\subsection{Framework}
\label{Framework}
In order to complete chemical tasks, we divided multi-agent into two types: agents with tools and agents without tools, and tested them on the first task, the Text-based Molecule Design task. The overall of our multi-agent framework is a modification of the framework of ~\citep{zhang2024cutcrapeconomicalcommunication} and ~\citep{qian2024scalinglargelanguagemodelbasedmultiagentcollaboration} that utilized different spatial and temporal masks to complete in the following six multi-agent structures: Chain, Random, FullConnected, Layered, Star and Debate mode. \\
However, since we are modifying their approach with a greater focus on information transmission and are also limited by API calls, we can only make a one-sided comparison regarding the recording of tokens and time.
\subsection{Implementation Details}
In multi-agent systems, information transmission is a critical factor for enhancing performance. By utilizing various structures, information can be conveyed through multiple pathways. To improve the efficiency of information collection, we have adopted the Final decision approach. Specifically, at the end of all structures, we have integrated a FinalRefer Agent to perform the final summary and decision-making. The FinalRefer prompt is followed:
\begin{tcolorbox}[colback=gray!10, colframe=black, title=Prompt: FinalRefer]
You are a strategic planning and final integration agent. You will be given a graduate-level question and reasoning outputs from all other agents.
Your task is to integrate all the information into a single, cohesive answer with detailed reasoning and evidence.\\
Your final output should:
1. Summarize the contributions from all agents, highlighting key insights.\\
3. Provide the final answer with a clear and detailed explanation.\\
4. Conclude with the final answer on a new line with the format: "The final answer is 'SMILES'\\
Here is the question:{question}. At the same time, the output of other agents is as follows:\\
{answers}
\end{tcolorbox}

In the implementation of the agents with tools, we modified all agents along the path except for the Final agent, while still following the ReAct framework for tool calling. During this process, due to the constraints of API calls, both the time required and the number of tokens used will be greater compared to agents without tools.
\subsection{Spatial Communication Topologies}
\subsubsection{Chain}
\begin{figure}[!h] 
    \centering
    \includegraphics[width=0.35\textwidth]{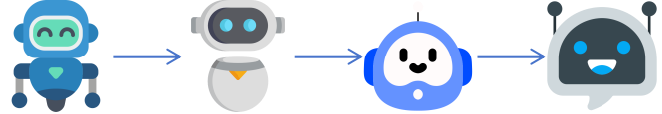}
    \captionsetup{font={small}} 
    \caption{
    Demonstration of \textbf{chain} structure}
    \label{fig:chain}
\end{figure}

The chain graph (Fig.~\ref{fig:chain}) is one of the most widely utilized communication architectures in contemporary multi-agent systems. In this architecture, the first agent receives input from the user, transforms it into new instruction, and subsequently forwards it to the next agent. Generally, the final agent in the chain provides a summary and answers.
\subsubsection{Random}
\begin{figure}[!h] 
    \centering
    \includegraphics[width=0.19\textwidth]{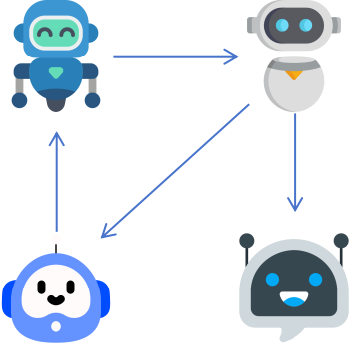}
    \captionsetup{font={small}} 
    \caption{
    Demonstration of \textbf{random} structure}
    \label{fig:random}
\end{figure}

The random graph refers to a sparse graph randomly sampled from a complete graph, as shown in the Fig.~\ref{fig:random}. They will execute asynchronously in multiple rounds and then randomly transmit information to the target agent. Finally, all the answers and information will transmit to the Final agent to make a final answer.
\subsubsection{FullConnected}
\begin{figure}[!h] 
    \centering
    \includegraphics[width=0.29\textwidth]{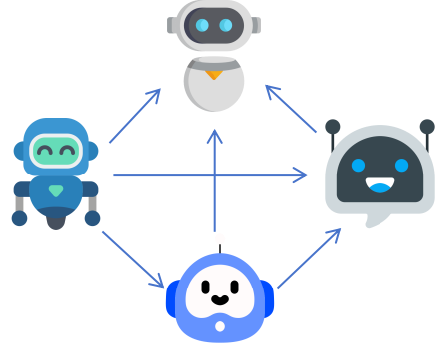}
    \captionsetup{font={small}} 
    \caption{
    Demonstration of \textbf{FullConnected} structure}
    \label{fig:fullconnected}
\end{figure}

The fullconnected graph (Fig.~\ref{fig:fullconnected}) is a directed graphs compared to traditional fully linked undirected graphs, which transmit information in a certain order to complete this topology structure. The final agent summarizes the dialogue and provides a concluding output or reflection.
\subsubsection{Layered}
\begin{figure}[!h] 
    \centering
    \includegraphics[width=0.35\textwidth]{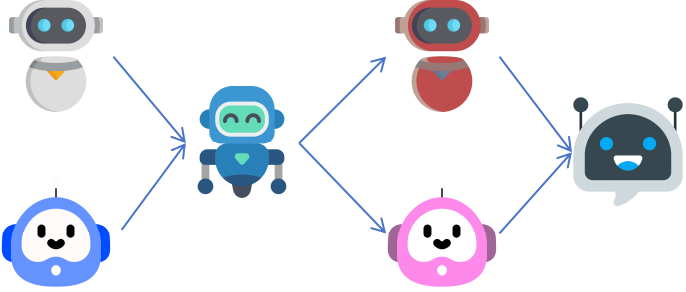}
    \captionsetup{font={small}} 
    \caption{
    Demonstration of \textbf{Layered} structure}
    \label{fig:Layered}
\end{figure}

The layered graph (Fig.~\ref{fig:Layered},~\citep{qian2024scalinglargelanguagemodelbasedmultiagentcollaboration}) refers to a stacked configuration similar to a multilayer perceptron (MLP). The first layer agents will feed to the agents in the second layer, and the final layer will make the summary and final-decision.
\subsubsection{Star}
\begin{figure}[!h] 
    \centering
    \includegraphics[width=0.23\textwidth]{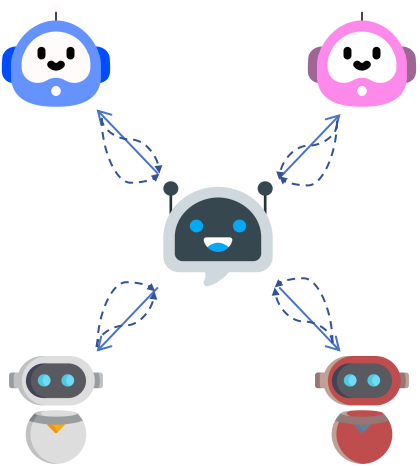}
    \captionsetup{font={small}} 
    \caption{
    Demonstration of \textbf{Star} structure}
    \label{fig:Star}
\end{figure}

The star graph (Fig.~\ref{fig:Star}) resembles the tree structure. Firstly, the problem will be handed over to the external leaf nodes for processing, and the obtained answer will be passed to the central root node, which will be repeated multiple times. Finally, the root node will give a summary and make the decision.
\subsubsection{Debate}
\begin{figure}[!h] 
    \centering
    \includegraphics[width=0.29\textwidth]{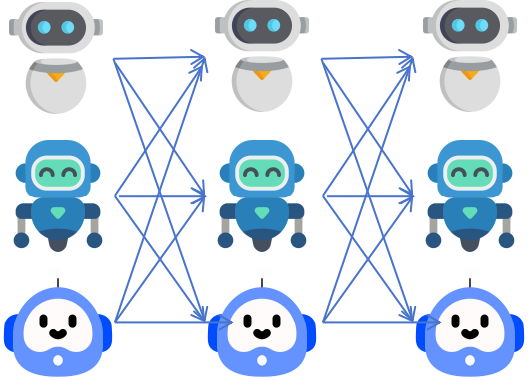}
    \captionsetup{font={small}} 
    \caption{
    Demonstration of \textbf{Debate} structure}
    \label{fig:Debate}
\end{figure}

The debate graph (Fig.~\ref{fig:Debate},~\citep{du2023improvingfactualityreasoninglanguage}) is multiple agents to engage in a debate, where in each round, every agent receives the outputs of all agents from the previous round before making their own statements. Generally, the finalRefer agent will help them to make the final decision.

\subsubsection{Tree-like\&Ours}
\begin{figure}[!h] 
    \centering
    \includegraphics[width=0.5\textwidth]{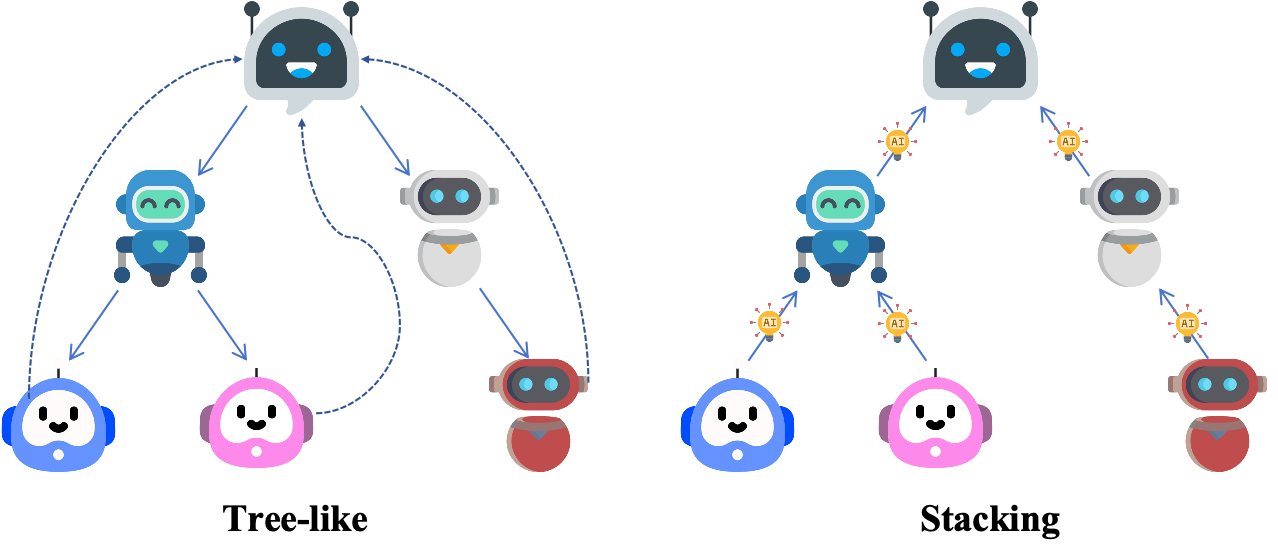}
    \captionsetup{font={small}} 
    \caption{
    Demonstration of \textbf{Tree} and \textbf{Ours} structure. In the composite tool structure, the icon '\includegraphics[height=1em]{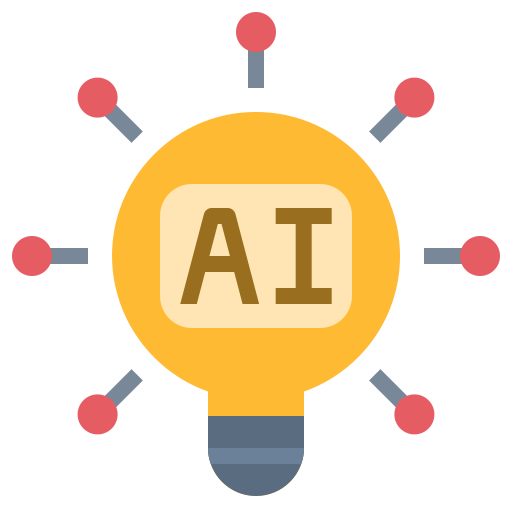}' means the parent node can choose which child node's information to accept.}
    \label{fig:Stacking}
\end{figure}

In order to better understand our composite tool structure, we will compare it with a tree like multi-agent pipeline that is relatively similar. The tree graph usually has the root node as the manager to supervise the nodes below to complete various tasks, and finally return the results of the leaf nodes to the root node for processing. Overall, this is just a top-down process of information transmission. As showin in the Fig.~\ref{fig:Stacking}\\
As for our composite tool structure, from the perspective of information transmission, the main agent of the root node also receives the information completed from below, but there is a difference. For the so-called tool node, it is a bottom-up process. After continuous information superposition step by step, it is passed upward through the parent node and then given to the main agent for processing. It can be seen from the figure that in each transmission process, whether it is the root node or the parent node, they all selectively accept the information from the child node, and it is not like a tree structure that is passed downward.

\input{tables/case}

\section{Case Study}
\label{casestudy}
There are four distinct composite tool behavior patterns-\textbf{correct}, \textbf{modify}, \textbf{judge} and \textbf{reserve} that make the composite tool works. Here are three other  cases for these patterns in the Tab.~\ref{tab:modify}:

\subsection{Correct}
In the text-based molecular design task, the model uses two callable tools. Initially, Tool A predicts the SMILES string "C1(C)CC1" for cyclopropane, which is incorrect due to a misinterpretation of the three-membered ring, introducing an erroneous branch. To correct this, Tool B is invoked with the query "Cyclopropane" and returns the correct SMILES "C1CC1", consistent with the standard structure.
\subsection{Modify}
As shown in the table, this is a common approach to using various tools. First, the problem is decomposed, and the RAG tool is used to retrieve information on each sub-question to obtain a preliminary answer. Then, subsequent processing is carried out using computational tools. Alternatively, one can first obtain an answer through computational tools, then have the agent self-assess the correctness of that answer, and finally use the RAG tool for cross-verify, thereby refining the answer and improving accuracy.

\subsection{Judge}
Judge refers to the process of selecting between two candidate answers based on the model’s knowledge in chemistry. This usually happens when two agent tools are available. When confronted with two anonymous tools that have the same descriptions, the model often opts to call both tools simultaneously before making a judgment. Furthermore, when using GPT-4o as the agent model, it typically demonstrates excellent judgment abilities.
\subsection{Reserve}
Reserve is a relatively uncommon phenomenon for large language models. Due to the hallucination problem, they usually choose to cover up knowledge blind spots to generate answers. However, when faced with multiple tools, the agent faces several answers. After thinking, it does not judge which one is more suitable, so it chooses to reserve opinions. Although this will reduce accuracy, in some ways it is a good thing to emerge this ability.

%% file: tables/app1.tex
\setlength\tabcolsep{1pt}
\begin{table*}[!htb]
    \centering
    \captionsetup{font={small}} 

    \resizebox{0.93\textwidth}{!}{
    \begin{tabular}{|c|lc|c|}
\hline
\multirow{2}{*}{Task}                                                                                  & \multicolumn{2}{c|}{Validation}                                                                                                   & \multirow{2}{*}{Result} \\ \cline{2-3}
                                                                                                       & \multicolumn{1}{c|}{Final tool}                                                                                      & score &                         \\ \hline
\multirow{3}{*}{\begin{tabular}[c]{@{}c@{}}Text-based\\ Molecule\\ Design\end{tabular}}                & \multicolumn{1}{l|}{{[}'Name2SMILES\_3', 'ChemDFM\_0'{]}}                                                            & 0.80  & 0.90                    \\ \cline{2-4} 
                                                                                                       & \multicolumn{1}{l|}{\textbf{{[}{[}'ChemDFM\_0', 'Name2SMILES\_1'{]}, 'ChemDFM\_1'{]}}}                               & 0.81  & \textbf{0.93}           \\ \cline{2-4} 
                                                                                                       & \multicolumn{1}{l|}{{[}{[}'ChemDFM\_1','Name2SMILES\_1'{]},{[}'ChemDFM\_1','Name2SMILES\_2'{]}{]}}                   & 0.91  & 0.91                    \\ \hline
\multirow{3}{*}{\begin{tabular}[c]{@{}c@{}}Molecule\\ Captioning\end{tabular}}                         & \multicolumn{1}{l|}{\textbf{{[}'SMILES2Description\_2','TextChemT5\_0'{]}}}                                          & 0.79  & \textbf{0.73}           \\ \cline{2-4} 
                                                                                                       & \multicolumn{1}{l|}{{[}'SMILES2Description\_3'{]}}                                                                   & 0.70  & 0.65                    \\ \cline{2-4} 
                                                                                                       & \multicolumn{1}{l|}{{[}'TextChemT5\_1','SMILES2Description\_1'{]}}                                                   & 0.71  & 0.66                    \\ \hline
\multirow{3}{*}{\begin{tabular}[c]{@{}c@{}}Reaction\\ Prediction\end{tabular}}                         & \multicolumn{1}{l|}{\textbf{{[}'Chemformer\_0','SMILES2Property\_2'{]}}}                                             & 1.00  & \textbf{0.90}                    \\ \cline{2-4} 
                                                                                                       & \multicolumn{1}{l|}{{[}'Chemformer\_2'{]}}                                                                           & 0.90  & 0.89                    \\ \cline{2-4} 
                                                                                                       & \multicolumn{1}{l|}{{[}'Chemformer\_1','SMILES2Property\_1'{]}}                                                      & 0.90  & 0.85                    \\ \hline
\multirow{3}{*}{\begin{tabular}[c]{@{}c@{}}Property\\ Prediction\\ (BACE)\end{tabular}}    & \multicolumn{1}{l|}{{[}'UniMol\_1','SMILES2Property\_1'{]}}                                                          & 0.80  & 72.1                    \\ \cline{2-4} 
                                                                                                       & \multicolumn{1}{l|}{\textbf{{[}{[}'UniMol\_1', 'SMILES2Property\_0'{]}, 'SMILES2Property\_0'{]}}}                    & 0.82  & \textbf{81.4}                    \\ \cline{2-4} 
                                                                                                       & \multicolumn{1}{l|}{{[}'UniMol\_3'{]}}                                                                               & 0.75  & 78.6                    \\ \hline
\multirow{3}{*}{\begin{tabular}[c]{@{}c@{}}Property\\ Prediction\\ (BBBP)\end{tabular}}    & \multicolumn{1}{l|}{\textbf{{[}'UniMol\_1','SMILES2Property\_2'{]}}}                                                 & 0.68  & \textbf{71.1}           \\ \cline{2-4} 
                                                                                                       & \multicolumn{1}{l|}{{[}'SMILES2Property\_3'{]}}                                                                      & 0.75  & 69.3                    \\ \cline{2-4} 
                                                                                                       & \multicolumn{1}{l|}{{[}'UniMol\_2'{]}}                                                                               & 0.73  & 70.6                    \\ \hline
\multirow{3}{*}{\begin{tabular}[c]{@{}c@{}}Property\\ Prediction\\ (Clintox)\end{tabular}} & \multicolumn{1}{l|}{\textbf{{[}{[}'UniMol\_1', 'SMILES2Property\_0'{]}, {[}'UniMol\_1', 'SMILES2Property\_1'{]}{]}}} & 0.70  & \textbf{72.3}           \\ \cline{2-4} 
                                                                                                       & \multicolumn{1}{l|}{{[}'SMILES2Property\_2'{]}}                                                                      & 0.68  & 61.4                    \\ \cline{2-4} 
                                                                                                       & \multicolumn{1}{l|}{{[}'UniMol\_1','SMILES2Property\_1'{]}}                                                          & 0.65  & 69.1                    \\ \hline
\multirow{3}{*}{\begin{tabular}[c]{@{}c@{}}Property\\ Prediction\\ (HIV)\end{tabular}}     & \multicolumn{1}{l|}{\textbf{{[}{[}'SMILES2Property\_1', 'UniMol\_0'{]}, 'SMILES2Property\_1'{]}}}                    & 1.00  & \textbf{97.4}           \\ \cline{2-4} 
                                                                                                       & \multicolumn{1}{l|}{{[}'UniMol\_0', 'SMILES2Property\_1'{]}}                                                         & 0.85  & 90.1                    \\ \cline{2-4} 
                                                                                                       & \multicolumn{1}{l|}{{[}'UniMol\_2'{]}}                                                                               & 0.90  & 96.9                    \\ \hline
\multirow{3}{*}{\begin{tabular}[c]{@{}c@{}}Property\\ Prediction\\ (Tox21)\end{tabular}}   & \multicolumn{1}{l|}{\textbf{{[}'UniMol\_2'{]}}}                                                                      & 0.78  & \textbf{92.3}           \\ \cline{2-4} 
                                                                                                       & \multicolumn{1}{l|}{{[}'UniMol\_0','SMILES2Property\_2'{]}}                                                          & 0.80  & 79.6                    \\ \cline{2-4} 
                                                                                                       & \multicolumn{1}{l|}{{[}'SMILES2Property\_2'{]}}                                                                      & 0.85  & 74.9                    \\ \hline
\end{tabular}
}
    
    \caption{Stacking results of different tasks. The \textbf{bold} font represents the most suitable toolset obtained in the task experiment.}
    \label{tab:appendix_molecule_design}
\end{table*}

%% file: tables/data.tex
\setlength\tabcolsep{1pt}
\begin{table*}[!htb]
    \centering
    \tiny
    \definecolor{lightgray}{gray}{0.9} 
    \resizebox{0.98\textwidth}{!}{
    \begin{tabular}{llllcc} 
    \hline
    \textbf{Ability} & \textbf{Task}                    & \textbf{Task Type} & \textbf{Dataset}                     & \textbf{\#val} & \textbf{\#test}  \\ \hline
    Understanding    & Molecular Property Prediction    & Classification     & BBBP, HIV, BACE, Tox21, ClinTox      & 250              & 100               \\ \hline
    Reasoning        & Reaction Prediction              & Generation         & USPTO-MIT                            & 100              & 100               \\ \hline
     Reasoning & Text-Based Molecule Design             & Generation                  & ChEBI-20-MM                          & 100              & 100              \\ \hline
    Explaining       & Molecule Captioning              & Generation         & ChEBI-20-MM                          & 100              & 100              \\ \hline
    \end{tabular}
    }
    \caption{Details of the training and test sets for the four chemistry tasks.}
    \label{tab:dataset}
\end{table*}

%% file: tables/case.tex
\setlength\tabcolsep{1pt}
\begin{table*}[!htb]
    \centering
    \tiny
    \definecolor{lightgray}{gray}{0.9} 
    \resizebox{1\textwidth}{!}{
    \begin{tabular}[l]{|>{\raggedright\arraybackslash}m{1.5cm}|>{\raggedright\arraybackslash}m{4cm}|>{\raggedright\arraybackslash}m{10cm}|}
    \hline
    \textbf{Agent Behavior Pattern} & \textbf{Input} & \textbf{Output} \\ 
    \hline
    \textbf{Correct} 
    & The molecule is a cycloalkane composed of a three-carbon ring. It functions as an inhalation anesthetic. It is a cycloalkane and a member of the cyclopropane class. What is its predicted SMILES?
    \newline
    \newline
    \newline
    Gold Answer: C1CC1
    & According to the provided description, the molecule is cyclopropane, which is a cycloalkane formed by three carbon atoms in a ring, well known for its use as an inhalation anesthetic. To confirm the SMILES notation of cyclopropane, I used the tool-A.
    \newline
    \textbf{Action:} Call tool-A  
    \newline
    \textbf{Input:} What is the SMILES notation for cyclopropane (a three-carbon cycloalkane used as an inhalation anesthetic)?  
    \newline
    \textbf{Output:} CC1(C)CC1 
    \newline
    \textbf{Observation:} The returned "CC1(C)CC1" does not correspond to cyclopropane. Cyclopropane should be a simple three-carbon ring, typically represented as "C1CC1". Next, I will call the tool-B to further correct the result.
    \newline
    \textbf{Action:} Call tool-B  
    \newline
    \textbf{Input:} Cyclopropane  
    \newline
    \textbf{Output:} C1CC1  
    \newline
    \textbf{Observation:} Based on the known structure of cyclopropane, the result from this tool is correct. Therefore, I will provide the correct SMILES based on the understanding of the chemical structure.  
    \newline
    \textbf{Final Answer: C1CC1 }\\
    \hline
    \textbf{Modify} 
    & The molecule is a peptide anion that is the conjugate base of gamma-Glu-Abu, obtained by removal of protons from the two carboxy groups as well as protonation of the amino group; major species at pH 7.3. It is a conjugate base of a gamma-Glu-Abu. Please try to infer the SMILES of this molecule.
    \newline
    \newline
    \newline
    Gold Answer: CCC(NC(=O)CCC([NH3+])C(=O)[O-])C(=O)[O-]
    & To find the SMILES representation of the peptide anion that is the conjugate base of gamma-Glu-Abu, I first need to understand the structure of gamma-Glu-Abu. Gamma-Glu-Abu stands for gamma-glutamyl-aminobutyric acid. I will start by finding the SMILES of gamma-Glu-Abu.
    \newline
    \textbf{Action:} Call tool-B  
    \newline
    \textbf{Input:} gamma-Glu-Abu
    \newline
    \textbf{Output:} CCC(NC(=O)CCC(N)C(=O)O)C(=O)O 
    \newline
    \textbf{Observation:} The next step is to modify this SMILES to represent the conjugate base of gamma-Glu-Abu at pH 7.3, where the two carboxyl groups are deprotonated and the amino group is protonated. This involves changing the carboxyl groups from "-C(=O)O" to "-C(=O)[O-]" and the amino group from "-N" to "-[NH3+]".
    \newline
    \textbf{Action:} Call tool-A  
    \newline
    \textbf{Input:} Modify the "CCC(NC(=O)CCC([NH3+])C(=O)[O-])C(=O)[O-]
"  from "-C(=O)O" to "-C(=O)[O-]" and the amino group from "-N" to "-[NH3+].
    \newline
    \textbf{Output:} CCC(NC(=O)CCC([NH3+])C(=O)[O-])C(=O)[O-] 
    \newline
    \textbf{Thought:} I now know the final answer
    \newline
    \textbf{Final Answer: CCC(NC(=O)CCC([NH3+])C(=O)[O-])C(=O)[O-]}\\
    \hline
    \textbf{Judge}&The molecule is a cycloalkane composed of three carbon atoms to form a ring. It has a role as an inhalation anaesthetic. It is a cycloalkane and a member of cyclopropanes.Please try to infer the SMILES of this molecule.
    \newline
    \newline
    \newline
    Gold Answer: CC(=O)N[C@@H]1[C@@H](O[C@@H]2O[C
    @@H](C)[C@@H](O)[C@@H](O)[C@@H]2
    O)[C@H](O[C@@H]2O[C@H](CO)[C@H](O
    )[C@H](O)[C@H]2O[C@@H]2O[C@@H](C)
    [C@@H](O)[C@@H](O)[C@@H]2O)[C@@
    H](CO)O[C@H]1O
&The molecule described is a complex carbohydrate structure involving alpha-L-Fucp, beta-D-Galp, and D-GlcNAc units. The specific linkages and configurations are crucial for determining the correct structure. Given the complexity of the molecule, the best approach is to use the Molecule\_Design tools, which are designed to handle detailed questions about molecular structures. I will start with Molecule\_Design\_1 to see if it can provide a correct SMILES representation based on the detailed description provided.
\newline
\textbf{Action:} Call tool-C
\newline
\textbf{Action Input:} The molecule is an alpha-L-Fucp-(1->2)-beta-D-Galp-(1->3)-[alpha-L-Fucp-(1->4)]-D-GlcNAc where the glucosamine at the reducing end has beta-configuration at its anomeric centre. It has a role as an epitope. Please provide the SMILES of this molecule.
\newline
\textbf{Observation:}CC(=O)N[C@@H]1[C@@H](O[C@@H]2O[C@@H](C)[C@@H](O)[C@@H](O)[C@@H]2O)
[C@H]
(O[C@@H]2O[C@H](CO)[C@H](O)[C@H](O)[C@H]2O[C@@H]2O
[C@@H](C)[C@@H](O)[C@@H](O)[C@@
H]2O)[C@@H](CO)O[C@H]1O
\newline
\textbf{Thought:}The SMILES provided by tool-C seems to represent a complex carbohydrate structure, but to ensure accuracy, I will use tool-D for a second opinion. This will help verify the correctness of the SMILES or provide an alternative if there are discrepancies.
\newline
\textbf{Action:} Call tool-D
\newline
\textbf{Action Input:} The molecule is an alpha-L-Fucp-(1->2)-beta-D-Galp-(1->3)-[alpha-L-Fucp-(1->4)]-D-GlcNAc where the glucosamine at the reducing end has beta-configuration at its anomeric centre. It has a role as an epitope. Please provide the SMILES of this molecule.
\newline
\textbf{Observation:}Nc1ccn([C@@H]2O[C@H](COP(=O)([O-])[O-])[C@@H](O)[C@H]2O)c(=O)n1
\newline
\textbf{Thought:}The SMILES provided by tool-D is significantly different from the one provided by tool-C, indicating a discrepancy. The structure from tool-D seems less likely to represent the complex carbohydrate structure described in the question. The SMILES from tool-C appears more plausible as it includes multiple sugar units linked together, which aligns better with the description of the molecule involving alpha-L-Fucp, beta-D-Galp, and D-GlcNAc. Therefore, I will consider the SMILES from tool-C as the more likely correct representation.
\newline
\textbf{Thought:} I now know the final answer
\newline
\textbf{Final Answer: CC(=O)N[C@@H]1[C@@H](O[C@@H]2O[C@@H](C)[C@@H](O)[C@@H](O)[C@@H]2O)
[C@H](O[C@@H]2O[C@H](CO)[C@H](O)[C@H](O)[C@H]2O
[C@@H]2O[C@@H](C)[C@@H](O)[C@@H]
(O)[C@@H]2O)[C@@H](CO)O[C@H]1O\textcolor{red}{\ding{52}}}\\ \hline
\textbf{Reserve}&The molecule is a linear amino tetrasaccharide composed of N-acetyl-beta-D-galactosamine, alpha-D-galactose, beta-D-galactose and N-acetyl-beta-D-galactosamine units joined by sequential (1->3)-, (1->4)- and (1->4)-linkages. It has a role as an epitope.Please try to give SMILES of this molecule.
    \newline
    \newline
    \newline
    Gold Answer: CC(=O)N[C@@H]1[C@@H](O)[C@H](O[C@
    @H]2O[C@H](CO)[C@H](O[C@H]3O[C@H]
    (CO)[C@H](O)[C@H](O[C@@H]4O[C@H](C
    O)[C@H](O)[C@H](O)[C@H]4NC(C)=O)[C@
    H]3O)[C@H](O)[C@H]2O)[C@@H](CO)O[C
    @H]1O
&The SMILES representation for the molecule described is not readily available due to the complexity of the glycosidic linkages and the specific stereochemistry involved. Further detailed chemical synthesis knowledge or specialized software would be required to accurately construct the full SMILES notation of this tetrasaccharide.
\newline
\textbf{Thought:}The initial attempt using the Molecule\_Design\_2 tool did not yield a direct SMILES result, likely due to the complexity of the molecule described. As an alternative approach, I can attempt to construct the molecule step-by-step using the tool-B tool for each individual sugar unit and then logically deduce the possible structure of the entire molecule. However, constructing the full SMILES for the entire molecule including specific linkages manually might not be feasible without specialized chemical software. I will first retrieve the SMILES for each sugar unit.
\newline
\textbf{Action:} Call tool-B
\newline
\textbf{Action Input:} N-acetyl-beta-D-galactosamine
\newline
\textbf{Observation:} CC(=O)N[C@@H]1[C@@H](O)[C@@H](O)[C@@H](CO)O[C@H]1O
\newline
\textbf{Thought:}I have obtained the SMILES for N-acetyl-beta-D-galactosamine. Next, I will retrieve the SMILES for alpha-D-galactose.
\newline
\textbf{Action:} Call tool-B
\newline
\textbf{Action Input:} alpha-D-galactose
\newline
\textbf{Observation:}OC[C@H]1O[C@H](O)[C@H](O)[C@@H](O)[C@H]1O
\newline
\textbf{Thought:}I have obtained the SMILES for alpha-D-galactose. Next, I will retrieve the SMILES for beta-D-galactose.
\newline
\textbf{Action:} Call tool-A
\newline
\textbf{Action Input:} What is the SMILES of beta-D-galactose
\newline
\textbf{Observation:} OC[C@H]1O[C@@H](O)[C@H](O)[C@@H](O)[C@H]1O
\newline
\textbf{Thought:}I have obtained the SMILES for beta-D-galactose. Since the molecule starts and ends with N-acetyl-beta-D-galactosamine, and I already have the SMILES for this unit, I do not need to retrieve it again. However, constructing the full SMILES string for the tetrasaccharide with specific glycosidic linkages (1->3, 1->4, 1->4) between these units is beyond the capability of the Call tool-B tool and requires specialized chemical synthesis software or expertise in carbohydrate chemistry. Therefore, I will conclude my process.
\newline
\textbf{Thought: }I now know the final answer
\newline
\textbf{Final Answer: Unable to provide the exact SMILES string for the described tetrasaccharide due to the complexity of the glycosidic linkages and the limitations of the tools available.}\\ \hline
    \end{tabular}
    }
    \caption{Case analysis of some behavior patterns of agents.}
    \label{tab:modify}
\end{table*}